\documentclass[10pt,twocolumn,letterpaper]{article}

\usepackage[pagenumbers]{cvpr} 

\usepackage[dvipsnames]{xcolor}

\usepackage{multirow}
\usepackage{colortbl}
\usepackage{amssymb}

\definecolor{cvprblue}{rgb}{0.21,0.49,0.74}
\definecolor{Gray}{gray}{0.9}
\usepackage[pagebackref,breaklinks,colorlinks,citecolor=cvprblue]{hyperref}

\def\paperID{2} 
\def\confName{CVPR}
\def\confYear{2024}

\title{Scaling Graph Convolutions for Mobile Vision}

\author{William Avery\\
The University of Texas at Austin\\
{\tt\small williamaavery@utexas.edu}
\and
Mustafa Munir\\
The University of Texas at Austin\\
{\tt\small mmunir@utexas.edu}
\and
Radu Marculescu\\
The University of Texas at Austin\\
{\tt\small radum@utexas.edu} \\
}

\begin{document}
\maketitle

\begin{abstract}
To compete with existing mobile architectures, MobileViG introduces Sparse Vision Graph Attention (SVGA), a fast token-mixing operator based on the principles of GNNs. However, MobileViG scales poorly with model size, falling at most 1\% behind models with similar latency. This paper introduces Mobile Graph Convolution (MGC), a new vision graph neural network (ViG) module that solves this scaling problem. Our proposed mobile vision architecture, MobileViGv2, uses MGC to demonstrate the effectiveness of our approach. MGC improves on SVGA by increasing graph sparsity and introducing conditional positional encodings to the graph operation. Our smallest model, MobileViGv2-Ti, achieves a 77.7\% top-1 accuracy on ImageNet-1K, 2\% higher than MobileViG-Ti, with 0.9 ms inference latency on the iPhone 13 Mini NPU. Our largest model, MobileViGv2-B, achieves an 83.4\% top-1 accuracy, 0.8\% higher than MobileViG-B, with 2.7 ms inference latency. Besides image classification, we show that MobileViGv2 generalizes well to other tasks. For object detection and instance segmentation on MS COCO 2017, MobileViGv2-M outperforms MobileViG-M by 1.2 $AP^{box}$ and 0.7 $AP^{mask}$, and MobileViGv2-B outperforms MobileViG-B by 1.0 $AP^{box}$ and 0.7 $AP^{mask}$. For semantic segmentation on ADE20K, MobileViGv2-M achieves 42.9\% $mIoU$ and MobileViGv2-B achieves 44.3\% $mIoU$ \footnote{Code: \url{https://github.com/SLDGroup/MobileViGv2}}.
\end{abstract}

\begin{figure}[h]
\centering
\includegraphics[scale=0.4]{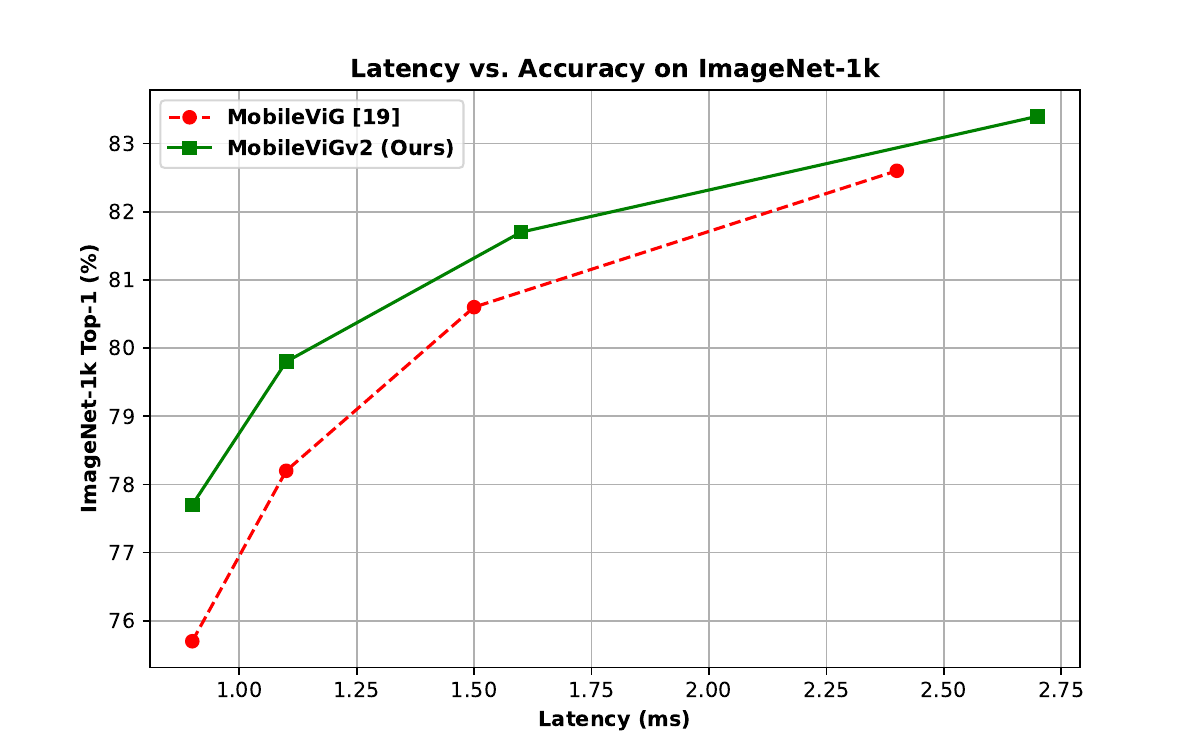}
\caption{Latency versus top-1 \% accuracy on ImageNet-1K of MobileViG \cite{MobileViG} and MobileViGv2. From this graph, we can see that MobileViGv2 improves on MobileViG, shifting the accuracy-latency curve up for similar points of inference latency.}
\label{fig:acclat}
\end{figure}

\section{Introduction}
\begin{figure*}[t]
\centering
\includegraphics[scale=0.9]{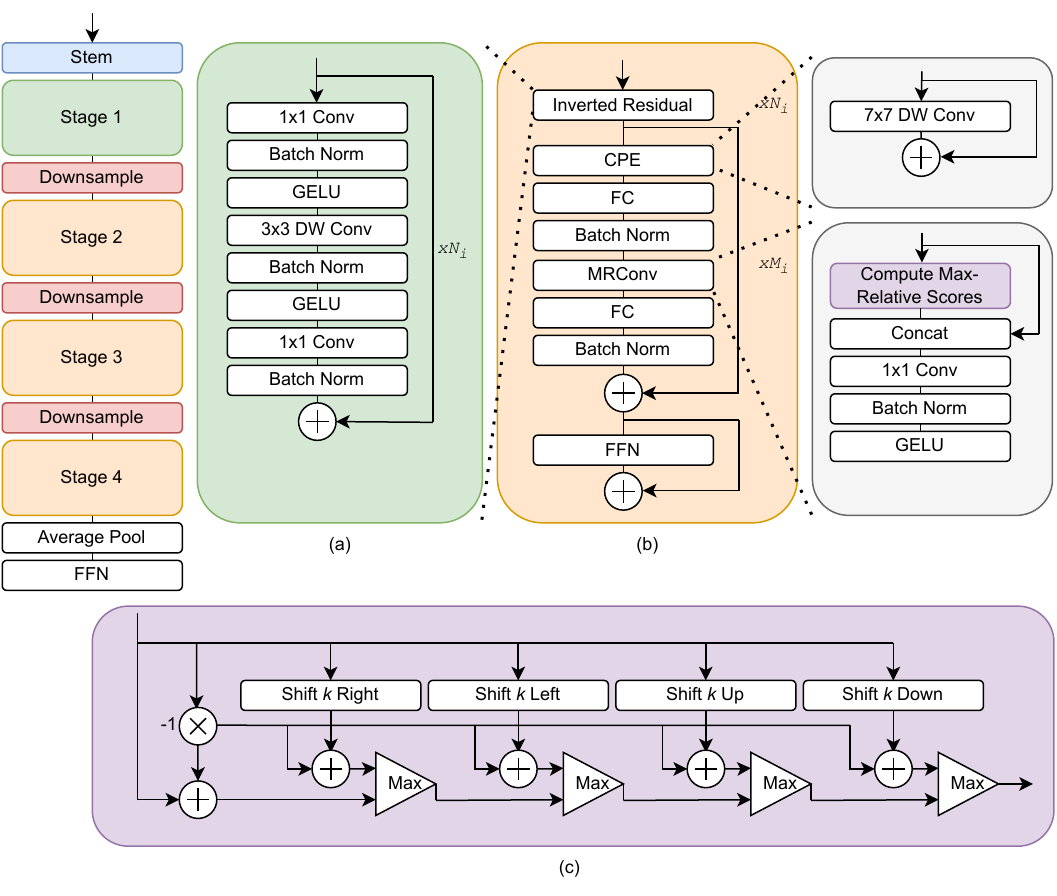}
\caption{The new MobileViGv2 architecture. The full architecture is shown on the left. The stem is composed of two stride two convolutions that downsample the input image by 4$\times$. Each downsampling block contains a single stride two convolution to downsample the input by 2$\times$. (a) An inverted residual block using GELU activation. For Stage 1, only inverted residuals are used. The number of inverted residuals in this stage is controlled by $N_{1}$. (b) For stages 2-4, a combination of inverted residuals and MGCs are used. Each stage has $N_{i}$ inverted residuals followed by $M_{i}$ MGCs, where $i$ is the stage number. The CPE block is a conditional positional encoding \cite{CPE} implemented with a 7$\times$7 depthwise convolution. The MRConv block contains graph construction and the max-relative message passing step. (c) Computing max-relative features using graph construction as outlined in MGC. Given an input image, this module computes the max-relative score against a fixed set of shifted inputs: shifting right, left, up, and down by $k$. The outputs of this stage are the max-relative scores, which are concatenated to the input and passed through a 1$\times$1 convolution to complete message passing.}
\label{fig:arch}
\end{figure*}

With the explosion of interest in large generative models, the demand for artificial intelligence (AI) applications has skyrocketed. An increasingly important sector of this demand is the mobile market. The goal is the ability to run powerful AI applications directly on user devices. These models must provide quick, personalized responses and, more importantly, keep users' data private and off the cloud. To achieve this, models must be small in size, fast, low power, and still maintain high performance on the target task.

Early efforts at targeting vision applications on mobile devices used convolutional neural networks (CNNs), such as with the MobileNet \cite{MobileNet}\cite{MobileNetv2} and EfficientNet\cite{tan2019efficientnet}\cite{tan2021efficientnetv2} family of architectures. While these models perform well, the introduction of vision transformers (ViTs) \cite{ViT} brought in new hybrid CNN-ViT mobile architectures \cite{MobileFormer}\cite{MobileViT}\cite{MobileViTv2}\cite{EfficientFormer}\cite{EfficientFormerv2} that significantly outperformed their CNN counterparts. The success of CNN-ViT-based mobile architectures over CNN-based ones is mainly due to the global receptive field of the self-attention operation, which accounts for more complex relationships across tokens. However, this success comes at a cost. The self-attention module is much slower than a convolution layer, as it scales quadratically with the number of input tokens. As such, most CNN-ViT-based mobile architectures only use self-attention in low-resolution stages.

More recently, vision graph neural networks (ViGs) \cite{Vision_GNN} were introduced as an alternative to CNNs and ViTs. ViGs connect tokens based on a predefined algorithm, such as K-nearest neighbors (KNN), and then mix the tokens with a message-passing scheme. The first use of ViGs in mobile vision architectures came with MobileViG \cite{MobileViG}. MobileViG uses Sparse Vision Graph Attention (SVGA), which replaces KNN with a static graph construction method, resulting in a fast CNN-ViG architecture. While MobileViG performs well for small model sizes, it scales poorly as the model size increases, falling nearly 1\% behind CNN-ViT-based models with similar latency \cite{EfficientFormerv2} \cite{FastViT}.

To address this scaling problem, we propose a new ViG module called \textit{Mobile Graph Convolution} (MGC), which improves on SVGA by increasing graph sparsity and introducing positional encodings to the graph operation. To demonstrate the effectiveness of MGC, we introduce MobileViGv2. This CNN-ViG-based architecture uses inverted residual blocks for processing using local receptive fields and MGC blocks for processing using long-range receptive fields. With higher graph sparsity, MobileViGv2 can use MGC at higher resolution stages without impacting latency compared to MobileViG. Unlike the original ViG \cite{Vision_GNN} model, MobileViG does not use positional encodings, an improvement introduced in MGC that leads to a significant performance boost with a slight increase in parameters. We summarize our contributions below:

\begin{enumerate}
    \item We propose Mobile Graph Convolution (MGC). This new mobile ViG module creates sparser graphs than Sparse Vision Graph Attention (SVGA) by fixing the number of possible connections per token regardless of input size. It uses conditional positional encodings to share the spatial relationships between tokens during message passing.
    \item We propose MobileViGv2, as shown in Figure \ref{fig:arch}, a CNN-ViG-based mobile architecture that uses MGC to achieve similar performance to state-of-the-art CNN-ViT-based mobile architectures. Notably, MobileViGv2 can begin mixing tokens globally at much higher resolution stages than existing CNN-ViT-based architectures due to the high speed of MGC.
    \item Our results show that a CNN-GNN-based mobile vision architecture can compete with state-of-the-art CNN-ViT-based image classification models and outperform them on downstream tasks. These results include latency and top-1 accuracy on ImageNet-1K \cite{imagenet1k}, object detection and instance segmentation on MS COCO 2017 \cite{coco}, and semantic segmentation on ADE20K \cite{ADE20K}.
\end{enumerate}

The remainder of this paper is structured as follows. Section 2 covers recent works in the mobile architecture space. Section 3 provides background on Sparse Vision Graph Attention (SVGA), the ViG module used in MobileViG \cite{MobileViG}. Section 4 describes the design of the MGC module and MobileViGv2 architecture. Section 5 describes the experimental setup and results for ImageNet-1K image classification, COCO object detection, COCO image segmentation, and ADE20K semantic segmentation. Additionally, Section 5 includes ablation studies further outlining the improvements of MGC and MobileViGv2 over SVGA and MobileViG. Section 6 summarizes our contributions.

\section{Related Work}
We break up previous works in the mobile vision space into three categories: CNN-based, CNN-ViT-based, and CNN-GNN-based. The most well known CNN-based methods are MobileNet \cite{MobileNet} \cite{MobileNetv2} \cite{MobileNetv3} and EfficientNet \cite{tan2019efficientnet} \cite{tan2021efficientnetv2}. MobileNet introduced depthwise separable convolutions, which separate the convolution operation into a depthwise convolution followed by a pointwise convolution. This approach achieves performance similar to a normal convolution op with significantly lower computational cost. MobileNetv2 \cite{MobileNetv2} built on depthwise separable convolutions with the new inverted residual block. Inverted residuals make residual links less memory-intensive on mobile devices by adding skip links around points where the channel layer is expanded. Hence, these large layers do not have to be saved in memory for future additions. Lastly, MobileNetv3 \cite{MobileNetv3} uses neural architecture search and squeeze and excitation blocks to further improve on MobileNetv2. Like MobileNetv3, EfficientNet \cite{tan2019efficientnet} and EfficientNetv2 \cite{tan2021efficientnetv2} use neural architecture search to produce fast, highly accurate models.

There are many CNN-ViT-based models, but two recent works have achieved remarkable performance with very low mobile latency: EfficientFormerV2 \cite{EfficientFormerv2} and FastViT \cite{FastViT}. EfficientFormerV2 combines inverted residual blocks with a modified transformer operation and SuperNet architecture search. The modified transformer block uses talking heads and a depthwise convolution on the value matrix to inject local information. FastViT \cite{FastViT} uses a new RepMixer block along with transformers to achieve similar results to that of EfficientFormerV2. The RepMixer block combines a depthwise convolution and convolution-based feed-forward network. Additionally, FastViT makes extensive use of reparamaterization.

To our knowledge, there is currently only one CNN-GNN-based mobile architecture: MobileViG \cite{MobileViG}. MobileViG introduced Sparse Vision Graph Attention (SVGA), a vision graph neural network module for statically constructing graphs and performing message passing. For small model sizes, MobileViG appears to compete with state-of-the-art CNN-ViT-based models, but as the model size grows, MobileViG accuracy fails to scale as well as that of CNN-ViT-based counterparts. To address this limitation, we introduce a new CNN-GNN-based mobile architecture, MobileViGv2, which fixes this scaling problem.

\section{Background}
Before explaining our proposed solution, Mobile Graph Convolution (MGC), to the scaling problem experienced by MobileViG \cite{MobileViG}, we briefly explain SVGA to highlight the differences between the two approaches better. SVGA can be split into a token mixing step and a feed-forward network. Given an input $X\in\mathbb{R}^{N \times M}$, 
\begin{equation} \label{eq:1}
\tag{1}
Y=\text{MRConv}(XW_{in})W_{out}+X
\end{equation}
where $Y\in\mathbb{R}^{N \times M}$ is the output of the token mixer, and $W_{in}$ and $W_{out}$ are fully connected layer weights.

For a single token $x_{i}$, the $MRConv$ operation \cite{maxrel} is described as follows,
\begin{equation} \label{eq:2}
\tag{2}
MRConv(x_{i}) = max(\{x_{j} - x_{i} | x_{j} \in G(x_{i})\})
\end{equation}
where $x_j$ is a neighboring token to $x_i$, in the set of neighboring tokens $G(x_{i})$.

The output of the token-mixing operation is then passed to a feed-forward network. The feed-forward network is a two-layer MLP expressed as
\begin{equation} \label{eq:3}
\tag{3}
Z=\sigma(XW_{1})W_{2}+Y
\end{equation}
where $Z\in\mathbb{R}^{N \times M}$, $W_{1}$ and $W_{2}$ are fully connected layer weights, and $\sigma$ is a GeLU activation.

\section{Methodology}
In this section, we describe the design of Mobile Graph Convolution (MGC) and MobileViGv2. MobileViGv2 leverages the speed of MGC to use graph convolutions in higher resolution stages of the model architecture, resulting in significantly higher model accuracy on ImageNet-1K without a significant drop in latency.

\begin{figure}[h]
\centering
\includegraphics[scale=0.675]{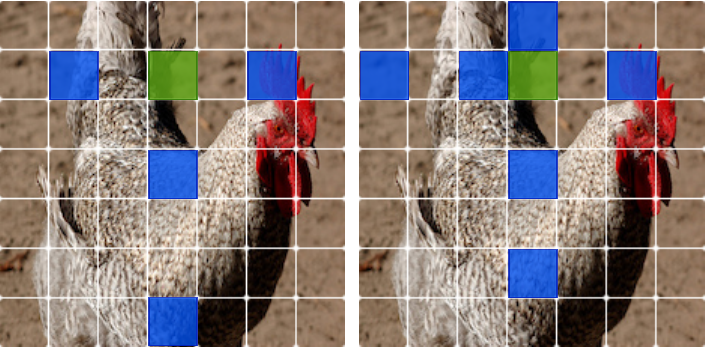}
\caption{Mobile Graph Convolution (MGC) (left) versus Sparse Vision Graph Attention (SVGA) (right). Each grid is broken up such that the effective receptive field is equal to that of MobileViGv2 at Stage 4. \textbf{(left)} The connections made for the green token using MGC ($L=2$) are shown in blue. \textbf{(right)} The connections made for the green token using SVGA ($K=2$), as used in MobileViG, are shown in blue. The image above was obtained from the ImageNet-1K \cite{imagenet1k} dataset and has been modified for this paper.}
\label{fig:mgcvssvga}
\end{figure}

\subsection{Mobile Graph Convolution}
We propose Mobile Graph Convolution (MGC) as a faster, highly scalable alternative to Sparse Vision Graph Attention (SVGA) \cite{MobileViG}. MGC improves on SVGA by increasing graph sparsity and introducing conditional positional encodings to the graph operation.

The MGC algorithm alters equations \ref{eq:1} and \ref{eq:2}, leaving \ref{eq:3} unchanged. Equation \ref{eq:1} is altered by adding a conditional positional encoding \cite{CPE} (CPE), and in equation \ref{eq:2}, the cardinality of $G(x_{i}) \forall x_{i}$ is reduced by using a different graph construction method. The updated equation for MGC is:
\begin{equation}
\tag{4}
Y=\text{MRConv}((X+CPE(X))W_{in})W_{out}+X
\end{equation}
where $CPE$ is a depthwise convolution and stands for conditional positional encoding.

\begin{table*}[ht]
\def\arraystretch{1.2}
\caption{Results of MobileViGv2 and other mobile architectures on ImageNet-1K classification task roughly grouped by NPU latency of an iPhone13 Mini using the ModelBench \cite{MobileOne} application. Type indicates whether the model is CNN-based, CNN-ViT-based, or CNN-GNN-based. Params lists the number of model parameters in millions. GMACs lists the number of MACs in billions. Gray highlights indicate the contributions of this paper. The Top-1 accuracy results for MobileViGv2 models are averaged over three experiments, and there is about a 0.1\% fluctuation between training seeds. Missing entries could not be profiled on the iPhone 13 Mini (iOS 16). $\downarrow$ means the lower, the better. $\uparrow$ means the higher, the better.}
\centering
\begin{tabular}[t]{c|c|c|c|c|c}
\hline
\multirow{2}{*}{\textbf{Model}} & \multirow{2}{*}{\textbf{Type}} & \multirow{2}{*}{\textbf{Params (M)}} & \multirow{2}{*}{\textbf{GMACs}} & \multirow{2}{*}{\textbf{iPhone 13 Latency $\downarrow$}} & \multirow{2}{*}{\textbf{Top-1 (\%) $\uparrow$}} \\
                                              &         &     &        & \textbf{(ms)} &      \\ \hline
MobileNetV2x1.0 \cite{MobileNetv2}            & CNN     & 3.5 & 0.3    & 0.8  & 71.8 \\
EdgeViT-XXS \cite{pan2022edgevits}            & CNN-ViT & 4.1 & 0.6    & -    & 74.4 \\
FastViT-T8  \cite{FastViT}                    & CNN-ViT & 3.6 & 0.7    & 0.8  & 76.7 \\
EfficientFormerV2-S0 \cite{EfficientFormerv2} & CNN-ViT & 3.5 & 0.4    & 0.8  & 75.7 \\
MobileViG-Ti \cite{MobileViG}                 & CNN-GNN & 5.2 & 0.7    & 0.9  & 75.7 \\
\rowcolor{Gray}
MobileViGv2-Ti                                & CNN-GNN & 5.6 & 0.6    & 0.9 & 77.7 \\ \hline

MobileNetV2x1.4 \cite{MobileNetv2}            & CNN     & 6.1 & 0.6    & 1.1   & 74.7 \\
EdgeViT-XS \cite{pan2022edgevits}             & CNN-ViT & 6.7 & 1.1    & -     & 77.5 \\
EfficientFormerV2-S1 \cite{EfficientFormerv2} & CNN-ViT & 6.1 & 0.7    & 1.0   & 79.0 \\
MobileViG-S \cite{MobileViG}                  & CNN-GNN & 7.2 & 1.0    & 1.1   & 78.2 \\
\rowcolor{Gray}
MobileViGv2-S                                 & CNN-GNN & 7.7 & 0.9    & 1.1 & 79.8 \\ \hline

EfficientNet-B0 \cite{tan2019efficientnet}    & CNN     & 5.3  & 0.4      & 1.5     & 77.7 \\
EdgeViT-S \cite{pan2022edgevits}              & CNN-ViT & 11.1 & 1.9      & -       & 81.0 \\
EfficientFormer-L1 \cite{EfficientFormer}     & CNN-ViT & 12.3 & 1.3      & 1.3     & 79.2 \\
FastViT-T12 \cite{FastViT}                    & CNN-ViT & 6.8  & 1.4      & 1.2     & 80.3 \\
FastViT-S12 \cite{FastViT}                    & CNN-ViT & 8.8  & 1.8      & 1.4     & 80.9 \\
FastViT-SA12 \cite{FastViT}                   & CNN-ViT & 10.9 & 1.9      & 1.6     & 81.9 \\
EfficientFormerV2-S2 \cite{EfficientFormerv2} & CNN-ViT & 12.6 & 1.3      & 1.5     & 81.6 \\
MobileViG-M \cite{MobileViG}                  & CNN-GNN & 14.0 & 1.5      & 1.5     & 80.6 \\
\rowcolor{Gray}
MobileViGv2-M                                 & CNN-GNN & 15.4 & 1.6      & 1.6       & 81.7 \\ \hline

EfficientNet-B3 \cite{tan2019efficientnet}    & CNN     & 12.2 & 2.0       & 4.8     & 81.6 \\
MobileViTv2-1.0 \cite{MobileViTv2}            & CNN-ViT & 4.9  & 1.8       & 3.0     & 78.1 \\
MobileViTv2-2.0 \cite{MobileViTv2}            & CNN-ViT & 18.5 & 7.5       & 6.3     & 82.4 \\
EfficientFormer-L3 \cite{EfficientFormer}     & CNN-ViT & 31.3 & 3.9       & 2.6     & 82.4 \\
EfficientFormer-L7 \cite{EfficientFormer}     & CNN-ViT & 82.1 & 10.2      & 6.5     & 83.3 \\
FastViT-SA24 \cite{FastViT}                   & CNN-ViT & 20.6 & 3.8       & 2.7     & 83.4 \\
EfficientFormerV2-L \cite{EfficientFormer}    & CNN-ViT & 26.1 & 2.6       & 2.4     & 83.3 \\
MobileViG-B \cite{MobileViG}                  & CNN-GNN & 26.7 & 2.8       & 2.4     & 82.6 \\
\rowcolor{Gray}
MobileViGv2-B                                 & CNN-GNN & 27.7 & 3.6       & 2.7     & 83.4 \\ \hline
\end{tabular}
\label{tab:class}
\end{table*}

Unlike the original ViG \cite{Vision_GNN}, MobileViG \cite{MobileViG} does not use positional encodings. As shown in ViT \cite{ViT}, without positional encodings, the model accuracy of ViTs drops significantly since the self-attention operation becomes permutation invariant. We find this performance drop also holds for graph operations in MobileViG \cite{MobileViG} (see Table \ref{tab:ablation}). As such, MGC uses conditional positional encodings (CPE) \cite{CPE} to encode spatial information before message passing. MGC uses reparameterizable CPE as introduced in FastViT \cite{FastViT}. Before constructing the graph and message passing, a positional encoding is added to the feature map by taking a depthwise convolution of the feature map itself. During inference, the residual link can be merged with the depthwise convolution, saving a fraction of a millisecond. This simple change introduces spatial information in the message-passing step yet adds few parameters and substantially increases performance. For example, when CPE \cite{CPE} is added to SVGA in MobileViG-B, the top-1 accuracy on ImageNet-1K improves by 0.3\%, as seen in Table \ref{tab:ablation}.

The differences in graph construction are straightforward. The SVGA algorithm uses a hyperparameter, $K$, to determine the distance and density of connections for each token. For token $x_{i}$, every $K^{th}$ token to the right in its row and down in its column is connected to it. Thus, the number of connections to $x_{i}$ grows according to $O(\frac{N+M}{K})$, where $N$ and $M$ are the dimensions of the input image. Thus, for a fixed value of $K$, the cardinality of $G(x_{i})$ grows linearly with respect to the input resolution. As such, using SVGA for higher resolutions requires more computation during graph construction, making it difficult to scale to higher input resolutions while maintaining competitive inference latency. One could fix this by scaling $K$ with input resolution, but a simple, more straightforward approach is to fix the number of possible connections per token regardless of input size.

Following this approach, each input token has only five connections in MGC: one self-connection, two long-range links to the left and right of the token on its row, and two long-range links up and down from the token on its column. This can be seen in greater detail in Figure \ref{fig:arch}c and Figure \ref{fig:mgcvssvga} for an input resolution of $7\times7$. For larger input resolutions, the distance of the long-range links can be increased to gather information from regions further away for each token. As implemented in MobileViG, SVGA uses $7$ connections per token ($K=2$), while MGC uses only $5$ ($L=2$), contributing to the speedup over SVGA. We also found that lowering the number of connections improves model performance, which can likely be attributed to reducing over-smoothing. For example, when swapping in SVGA for MGC in MobileViGv2-B, the resulting model is 0.2 milliseconds slower with 0.1\% worse top-1 classification accuracy on ImageNet-1K.

\begin{table*}[h]
\def\arraystretch{1.2}
\caption{Results of MobileViGv2 and other mobile architectures on COCO object detection, COCO instance segmentation tasks, and ADE20K semantic segmentation. Parameters lists the number of backbone parameters in millions, not including Mask-RCNN or Semantic FPN. $AP^{box}$ and $AP^{mask}$ scores are for object detection and instance segmentation on MS COCO 2017 \cite{coco}. $mIoU$ scores are for semantic segmentation on ADE20K \cite{ADE20K}. Shaded regions show the contributions of this paper. A (-) denotes a model that did not report these results.}
\centering
\begin{tabular}[t]{c|c|c|c|c|c|c|c|c}
\hline
\multirow{2}{*}{\textbf{Model}} & \multirow{2}{*}{\textbf{Params (M)}} & \multirow{2}{*}{$AP^{box}$} & \multirow{2}{*}{$AP^{box}_{50}$} & \multirow{2}{*}{$AP^{box}_{75}$} & \multirow{2}{*}{$AP^{mask}$} & \multirow{2}{*}{$AP^{mask}_{50}$} & \multirow{2}{*}{$AP^{mask}_{75}$} & \multirow{2}{*}{$mIoU$} \\
                                              & & & & & & & & \\ \hline

EfficientFormer-L1 \cite{EfficientFormer}     & 12.3 & 37.9 & 59.0 & 40.1 & 34.6 & 55.8 & 36.9 & 38.9 \\
FastViT-SA12 \cite{FastViT}                   & 10.9 & 38.9 & 60.5 & 42.2 & 35.9 & 57.6 & 38.1 & 38.0 \\
MobileViG-M \cite{MobileViG}                  & 14.0 & 41.3 & 62.8 & 45.1 & 38.1 & 60.1 & 40.8 & - \\
\rowcolor{Gray}
MobileViGv2-M                                 & 15.4 & \textbf{42.5} & \textbf{63.9} & \textbf{46.3} & \textbf{38.8} & \textbf{60.8} & \textbf{41.7} & \textbf{42.9}  \\ \hline

EfficientFormer-L3 \cite{EfficientFormer}     & 31.3 & 41.4 & 63.9 & 44.7 & 38.1 & 61.0 & 40.4 & 43.5 \\
FastViT-SA24 \cite{FastViT}                   & 20.8 & 42.0 & 63.5 & 45.8 & 38.0 & 60.5 & 40.5 & 41.0 \\
MobileViG-B \cite{MobileViG}                  & 26.7 & 42.0 & 64.3 & 46.0 & 38.9 & 61.4 & 41.6 & - \\
\rowcolor{Gray}
MobileViGv2-B                                 & 27.7 & \textbf{43.0} & \textbf{64.9} & \textbf{47.1} & \textbf{39.6} & \textbf{62.2} & \textbf{42.7} & \textbf{44.3}  \\ \hline
\end{tabular}
\label{tab:objinst}
\end{table*}

\subsection{MobileViGv2 Architecture}
The MobileViGv2 architecture, as shown in Figure \ref{fig:arch}, can be broken into four main stages, where processing occurs at a single resolution in a given stage. Within each stage, a mixture of inverted residuals and Mobile Graph Convolutions (MGCs) are used.

The entry point into the architecture is the stem. The stem takes the input image and downsamples it 4$\times$ using convolutions with stride equal to two. The output of the stem is fed to Stage 1, which consists of $N_{1}$ inverted residuals as described in Figure \ref{fig:arch}a. Between each stage is another convolution-based downsampling step. Stages 2, 3, and 4 each start with a sequence of $N_{i}$ inverted residuals, where $i$ is the stage number. The output of the inverted residual sequence is then fed through $M_{i}$ MGCs, as shown in Figure \ref{fig:arch}b. After Stage 4, an average pooling step followed by a feed-forward network produces the predicted class of the input image.

To achieve different model sizes, the channel width of each stage and values of $N_{i}$ and $M_{i}$ are changed. There are four different MobileViGv2 configurations, MobileViGv2-Ti, MobileViGv2-S, MobileViGv2-M, and MobileViGv2-B. Note that all inverted residual blocks use an expansion factor of 4. Additionally, all FFNs used in MGC, as shown in Figure \ref{fig:arch}b, use an expansion factor of 4. While a different mixture of widths and expansion factors may produce better results, as could be found using a neural-architecture search, our work aims to show the potential of using graph convolutions in mobile vision architectures, not finding the optimal model structure. We leave this task of using NAS for future work.

\begin{table*}[h]
\def\arraystretch{1.2}
\caption{An ablation study of the effects of conditional positional encodings, higher resolution graphers, and different graph construction methods on MobileViG-B and MobileViGv2-B. A checkmark indicates this component was used in the experiment. A (-) indicates this component was not used. 1-Stage indicates that graph convolutions were only used in Stage 4 of the model, while 3-Stage indicates that graph convolutions were used in stages 2, 3, and 4. The * indicates that this model swapped the graph convolution for a fully connected layer.}
\centering
\begin{tabular}[t]{c|c|c|c|c|c|c|c|c}
\hline
\multirow{2}{*}{\textbf{Base Model}} & \multirow{2}{*}{\textbf{Params (M)}} & \multirow{2}{*}{\textbf{Latency (ms)}} & \multirow{2}{*}{\textbf{SVGA}} & \multirow{2}{*}{\textbf{MGC}} & \multirow{2}{*}{\textbf{CPE}} & \multirow{2}{*}{\textbf{1-Stage}} & \multirow{2}{*}{\textbf{3-Stage}} & \multirow{2}{*}{\textbf{Top-1 (\%)}} \\
                                              &         &     &     &  &    &  &  \\ \hline

MobileViG-B                     & 26.7 & 2.4 & \checkmark & - & - & \checkmark & - & 82.6 \\
MobileViG-B                     & 27.6 & 2.9 & \checkmark & - & - & - & \checkmark & 83.2 \\
MobileViG-B                     & 26.8 & 2.4 & \checkmark & - & \checkmark & \checkmark & - & 82.9 \\
MobileViG-B                     & 27.7 & 2.9 & \checkmark & - & \checkmark & - & \checkmark & 83.3 \\
MobileViG-B*                    & 28.6 & 2.5 & - & - & - & - & \checkmark & 83.0 \\
\rowcolor{Gray}
MobileViGv2-B                   & 27.7 & 2.7 & -    & \checkmark  & \checkmark    & - & \checkmark & \textbf{83.4}   \\ \hline
\end{tabular}
\label{tab:ablation}
\end{table*}

\section{Experimental Results}
In this section, we describe the setup and results for MobileViGv2 experiments on ImageNet-1K \cite{imagenet1k} classification, COCO object detection, COCO instance segmentation \cite{coco}, and ADE20K \cite{ADE20K} semantic segmentation tasks. 

\subsection{Image Classification}

MobileViGv2 is implemented using PyTorch 1.12.1 \cite{paszke2019pytorch} and the Timm library \cite{timm}. Each model is trained using 16 NVIDIA A100 GPUs with an effective batch size of 2048. The models are trained from scratch for 300 epochs on the ImageNet-1K dataset with a standard training and inference resolution of 224$\times$224. We use the AdamW \cite{AdamW} optimizer and a learning rate of 2e-3 with a cosine annealing schedule. Like many CNN-ViT-based \cite{EfficientFormerv2} \cite{MobileViG} mobile architectures, we use RegNetY-16GF \cite{RegNetY} for knowledge distillation. Our data augmentation pipeline includes RandAugment \cite{RandAugment}, Mixup \cite{Mixup}, Cutmix \cite{CutMix}, random erasing \cite{RandomErase}, and repeated augment \cite{RepeatedAugment}. To measure inference latency, all models are packaged as MLModels using CoreML and profiled on the same iPhone 13 Mini (iOS 16) using ModelBench \cite{MobileOne}. We use the following ModelBench settings to profile each model: 50 inference rounds, 50 inferences per round, and a low/high trim of 10. Table \ref{tab:class} shows ImageNet-1K classification results for MobileViGv2 and similar mobile vision architectures. Models are roughly grouped by latency.

For models with an inference latency under 1 ms, MobileViGv2-Ti has the highest accuracy, with the next closest model, FastViT-T8 \cite{FastViT}, being a full 1\% behind. Additionally, MobileViGv2-Ti is 2\% more accurate than MobileViG-Ti for the same inference latency. When compared to EfficientFormerV2-S1 \cite{EfficientFormerv2}, MobileViGv2-S has 0.8\% higher accuracy for a similar inference latency. MobileViGv2-S is also 1.7\% more accurate than MobileViG-S for the same inference latency.

The wide performance gap shown in Figure \ref{fig:acclat} indicates that  MGC and the new model configuration of MobileViGv2 successfully solve the scaling problem experienced by MobileViG. This, along with better or comparable performance to CNN-ViT-based models such as EfficientFormerV2 \cite{EfficientFormerv2} and FastViT \cite{FastViT}, show the potential of CNN-GNN-based architectures to compete in the mobile vision space.

\subsection{Object Detection and Instance Segmentation}
We show that MobileViGv2 generalizes well to downstream tasks by using it as a backbone for object detection and instance segmentation on the MS COCO 2017 \cite{coco} dataset, which contains training and validation sets of 118K and 5K images. We use pre-trained MobileViGv2 backbones with Mask-RCNN \cite{mask_r_cnn} for training. Each model is trained on 16 NVIDIA A100 GPUs for 12 epochs with an effective batch size of 16. We use AdamW \cite{AdamW} optimizer, an initial learning rate of 2e-4, and a standard image resolution of 1333$\times$800.

As shown in Table \ref{tab:objinst}, MobileViGv2-M hits an $AP^{box}$ and $AP^{mask}$ of 42.5 and 38.8, respectively. This is 1.2 and 0.7 points higher than MobileViG-M \cite{MobileViG} and 3.6 and 2.9 points higher than FastViT-SA12 \cite{FastViT}. MobileViGv2-B achieves an $AP^{box}$ and $AP^{mask}$ of 43.0 and 39.6, respectively. This is 1.0 and 0.7 points higher than MobileViG-B and 1.0 and 1.6 points higher than the comparable FastViT backbone, FastViT-SA24.

These results show that MobileViGv2 generalizes well to downstream tasks. Compared to competitive CNN-ViT-based models like FastViT \cite{FastViT}, MobileViGv2 performs significantly better on these downstream tasks, even though the performance in image classification is comparable.

\subsection{Semantic Segmentation}
We also show that MobileViGv2 generalizes well to semantic segmentation on the ADE20K dataset \cite{ADE20K}, which contains 20K training images and 2K validation images with 150 semantic categories. For training, we use 8 NVIDIA RTX 6000 Ada generation GPUs, the AdamW optimizer, and a learning rate of 2e-4 with polynomial decay. We use MobileViGv2 as a backbone with Semantic FPN \cite{SemanticFPN} as the segmentation decoder. The backbone is initialized with pre-trained weights on ImageNet-1K, and the model is trained for 40K iterations.

As shown in Table \ref{tab:objinst}, MobileViGv2-M outperforms FastViT-SA12 \cite{FastViT} by 4.9\% $mIoU$ and outperforms EfficientFormer-L1 \cite{EfficientFormer} by 4\% $mIoU$. Additionally, MobileViGv2-B outperforms FastViT-SA24 by 3.3\% $mIoU$ and outperforms EfficientFormer-L3 by 0.8\% $mIoU$. Again, when compared to competitive CNN-ViT-based models like FastViT \cite{FastViT} and EfficientFormer \cite{EfficientFormer}, MobileViGv2 performs significantly better on this downstream task even though the performance in image classification is comparable.

\subsection{Ablation Studies}

We perform ablation studies to show the benefits of MGC over SVGA and to demonstrate that graph convolutions provide benefits over a simple feed-forward network solution. A summary of these results can be found in Table \ref{tab:ablation}.

Starting with MobileViG-B as a base model, we try using SVGA-style graph convolutions in stages 2, 3, and 4 of the model while keeping the number of parameters the same. We adjust the number of blocks in each stage and the channel depth to keep the number of parameters similar. The resulting model achieves a top-1 accuracy on ImageNet-1K of 83.2\%, 0.6\% higher than MobileViG-B. However, this model has an inference latency of 2.9 milliseconds, significantly slower than the 2.3 milliseconds of MobileViG-B without catching up to the top-1 performance of FastViT \cite{FastViT} and EfficientFormerV2 \cite{EfficientFormerv2}.

We also try adding CPE to SVGA in MobileViG-B and find that it improves model performance by 0.3\%. We then combine both CPE and 3-stage SVGA, which results in a top-1 accuracy of 83.3\%. Even though SVGA makes more connections than MGC, it performs slightly worse than MGC when used with the same model settings. We expect that this occurs due to over-smoothing from the higher connection count.

To get the benefits of using more stages without a significant hit to latency, we use MGC, which uses sparser graphs and CPE, in MobileViGv2-B, to achieve the best performance of 83.4\%. This final configuration has the same top-1 performance and mobile latency as FastViT-SA24 \cite{FastViT}.

To verify that graph convolutions boost model performance, we swap each graph convolution with a fully connected layer of the same expansion size and use this in stages 2, 3, and 4 of the model. This experiment is marked as MobileViG-B* in Table \ref{tab:ablation}. We find that this model achieves an accuracy of only 83.0\%, which is 0.2\% less than using SVGA in three stages and 0.4\% less than using MGC in three stages. This shows that graph convolutions are improving model performance.


\section{Conclusion}

In this work, we have proposed Mobile Graph Convolution (MGC) and MobileViGv2, a model architecture that uses MGC and competes with state-of-the-art CNN-ViT-based mobile architectures. MGC uses a sparser, static graph construction method than SVGA, resulting in faster inference speeds. Additionally, MGC introduces conditional positional encodings to the graph operation, considerably boosting model accuracy with only a slight increase in the number of parameters. With these changes, MGC can be used in much higher resolution stages than SVGA without significantly impacting latency. 

MobileViGv2 takes advantage of this by using MGC in the last three processing stages. Earlier global processing and the sharing of spatial information during message passing through CPE solves the scaling problem experienced by MobileViG, thus making MobileViGv2 a genuine competitor to state-of-the-art CNN-ViT-based mobile architectures and, consequently, MGC a competitor to self-attention in the mobile vision model space.

{
    \small
    \bibliographystyle{ieeenat_fullname}
    \bibliography{main}
}

\end{document}